\documentclass[10pt,conference]{IEEEtran}
\IEEEoverridecommandlockouts
\usepackage{graphicx}
\usepackage{subfig}
\usepackage{cite}
\usepackage{amsmath,amssymb,amsfonts}
\usepackage{algorithmic}
\usepackage{graphicx}
\usepackage{textcomp}
\usepackage{xcolor}
\usepackage{multirow}
\usepackage{array, makecell}
\usepackage{adjustbox}

\makeatletter
\renewcommand\paragraph{\@startsection{paragraph}{4}{\z@}%
	{1.5ex plus .2ex minus .3ex}%
	{-0em}%
	{\normalsize\bf}}
\makeatother

\def\BibTeX{{\rm B\kern-.05em{\sc i\kern-.025em b}\kern-.08em
    T\kern-.1667em\lower.7ex\hbox{E}\kern-.125emX}}

\begin{document}

\title{\huge DeepScale: Online Frame Size Adaptation for Multi-object Tracking on Smart Cameras and Edge Servers}
\author{Keivan Nalaie\\
McMaster University\\
{\tt\small nalaiek@mcmaster.ca}

\and
Renjie Xu\\
McMaster University\\
{\tt\small xur86@mcmaster.ca}
\and
Rong Zheng\\
McMaster University\\
{\tt\small rzheng@mcmaster.ca}
}

\maketitle
\thispagestyle{plain}
\pagestyle{plain}

\begin{abstract}
In surveillance and search and rescue applications, it is important to perform multi-target tracking (MOT) in real-time on low-end devices. Today's MOT solutions employ deep neural networks, which tend to have high computation complexity. Recognizing the effects of frame sizes on tracking performance, we propose DeepScale, a model agnostic frame size selection approach that operates on top of existing fully convolutional network-based trackers to accelerate tracking throughput. In the training stage, we incorporate detectability scores into a one-shot tracker architecture so that DeepScale can learn representation estimations for different frame sizes in a self-supervised manner. During inference, it can adapt frame sizes according to the complexity of visual contents based on user-controlled parameters. To leverage computation resources on edge servers,  we propose two computation partition schemes tailored for MOT, namely, edge server only with adaptive frame-size transmission and edge server-assisted tracking. Extensive experiments and benchmark tests on MOT datasets demonstrate the effectiveness and flexibility of DeepScale. Compared to a state-of-the-art tracker, DeepScale++, a variant of DeepScale achieves 1.57X accelerated with only moderate degradation ($\sim$ 2.3\%) in tracking accuracy on the MOT15 dataset in one configuration. We have implemented and evaluated DeepScale++ and the proposed computation partition schemes on a small-scale testbed consisting of an NVIDIA Jetson TX2 board and a GPU server. The experiments reveal non-trivial trade-offs between tracking performance and latency compared to server-only or smart camera-only solutions.
\end{abstract}
\section{Introduction}

Multi-Object Tracking (MOT) aims to detect and track the trajectories of moving objects in a visual scene. It is a key building block in live video analytics and can find applications in surveillance, search and rescue, and autonomous driving applications~\cite{wu20213d},~\cite{tokmakov2021learning}. Modern MOT systems~\cite{zhang2020fairmot},~\cite{centerTrack},~\cite{jde} typically adopt a {tracking-by-detection approach} where the bounding boxes of objects of interest in each frame are first identified via object detection, and then object association is applied to link the corresponding identities over the time based on appearance and/or motion features. 
Increasingly, deep neural networks (DNN), specifically convolutional neural networks (CNN) become the dominant models for object detection and re-identification in MOT~\cite{pang2020tubetk},~\cite{deepsort},~\cite{trackingwithoutshistles}. Despite their superior performance, these models tend to have high computation complexity making them inadequate for real-time applications at high input resolutions and frame rates on low-end devices such as smart cameras.  

To enable real-time prediction of DNN models, there are two complementary categories of approaches. First, one can deploy edge servers with high computation power close to the cameras and offload part of or all of their computation tasks. For example, several computation partition approaches have been proposed that split the data flow graph representation of a {\it given} model into two or more parts and execute them on end devices and edge servers~\cite{kang2017neurosurgeon,ko2018edge}. The second line of approaches reduce the size of DNN models through techniques such as model compression~\cite{yin2020dreaming,chen2015net2net,hinton2015distilling} and neural architecture search~\cite{nakandala2020cerebro,liu2018demand}. While both categories of solutions can be applied to MOT, they either fail to take advantage of inherent temporal and spatial redundancy in video streams or require a significant reworking of existing models. In this paper, we take a {\it model-agnostic} approach for real-time MOT and develop {\it DeepScale}, a frame resolution adaptation framework that scales input frames based on the complexity of their visual content. DeepScale is motivated by two key observations of MOT applications. First, higher resolution frames can capture fine details of remote objects in a scene and thus lead to more accurate tracking. However, not all scenes are created equal or contain relevant objects in far fields. Second, although key layers in CNN models such as convolutional layers and max-pooling layers can work with different input sizes without any modification, the inference time increases with increasing input frame sizes.  
From the two insights, we ask {\it i) is it possible to find the minimal frame sizes with comparable tracking accuracy as that of fixed high-resolution inputs; ii) can we provide tuning knobs to developers and end-users to control the trade-offs between computing time and tracking accuracy by adapting the frame size automatically? and iii) how to partition a DNN-based MOT pipeline between end devices and edge servers?}

The answers to the first two questions are affirmative. 
At run time, DeepScale takes every $K$th full-resolution frame as its input to estimate a suitable resolution for subsequent $K-1$ frames based on user-provided control parameters. Object detection and association are then performed on the scaled frames using {unmodified fully convolutional detector model architectures} from existing trackers.  To further reduce computation overheads, resolution estimation in DeepScale shares common feature extraction layers as object detection. Training of DeepScale is self-supervised. No extra label is needed other than those for the MOT task themselves. Instead, a detectability measure is introduced to gauge the ability of frames of different resolutions to capture relevant objects. 
To answer the third question, we propose two computation partition schemes tailored for MOT, namely, edge server-only with adaptive frame-size transmission (SOAT) and edge server-assisted tracking (SAT). Both schemes take advantage of the reduced frame resolutions offered by DeepScale to decrease the amount of network data transfer and computation loads on end devices. Moreover, the model-agnostic nature of DeepScale makes it possible to run models of different complexity on end devices and edge servers to fine-tune the trade-off between tracking accuracy and latency. 

To evaluate the performance of DeepScale, we conduct experiments on both public MOT datasets and a small-scale testbed consisting of an NVIDIA Jetson TX2 powered device and a Telsa P100 GPU server. 
DeepScale can achieve a good trade-off between computation time and tracking performance. For example, in comparison to a state-of-the-art (SOTA) method, DeepScale is able to accelerate the end-to-end inference speed up to 60\% at only 2.3\% reduction on multiple object tracking accuracy (MOTA) scores. 

In summary, the main contributions of the work include:
\begin{itemize}
\item We provide a quantitative study of the impact of frame sizes on tracking time and accuracy. 
\item We propose a model-agnostic DeepScale framework that adapts input frame sizes based on the visual content of scenes. 
\item An extension of DeepScale, called DeepScale++,  is further devised to take advantage of training with multi-resolution inputs to improve the inference accuracy of low-resolution frames.
\item Two computation partition approaches that split MOT inference tasks between an end device and an edge server are proposed and evaluated using a real-world testbed. 
\item Extensive experiments on MOT datasets have been conducted and demonstrate the superior performance and flexibility of DeepScale++ in accelerating MOT tasks. 
\end{itemize}

The rest of the paper is organized as follows. In Section \ref{impacts_of_frame_sizes_section}, we first quantify the impact of frame sizes on tracking time and accuracy and then present the training procedure and the inference pipeline of DeepScale. In Section \ref{experiments}, extensive experiment results on MOT datasets are presented by comparing our solution with other SOTA methods.
 In Section \ref{relatedWork}, we review prior works in accelerating video analytics tasks followed by a conclusion in Section \ref{conclusion}.

\section{Background and Motivation}
\subsection{Multi-object Tracking}
Visual multi-object tracking starts with detecting objects (using e.g.\cite{duan2019centernet},\cite{yolov3}) of interest in a frame. During tracking, a tracker (e.g. \cite{centerTrack},\cite{jde}) adds newly detected objects to an active set. In the case of pedestrian tracking, the tracker keeps a set of detected bounding boxes and an identity vector. For each person, bounding boxes with similar motion features in different frames are grouped into a track. Kalman filtering or its variants can be applied to predict the position of each track in the next frame.  Next, each detected object in the next frame is matched to an active track based on appearance features and the distance between the detected location and predicted position. If a track is matched to a detected object, the detection is considered as a part of the track. For unmatched detections, new tracks are created.
\subsection{Impacts of Frame Sizes on Tracking Performance}
\label{impacts_of_frame_sizes_section}
\begin{figure}[t]
\centering
\includegraphics[width=0.40\textwidth]{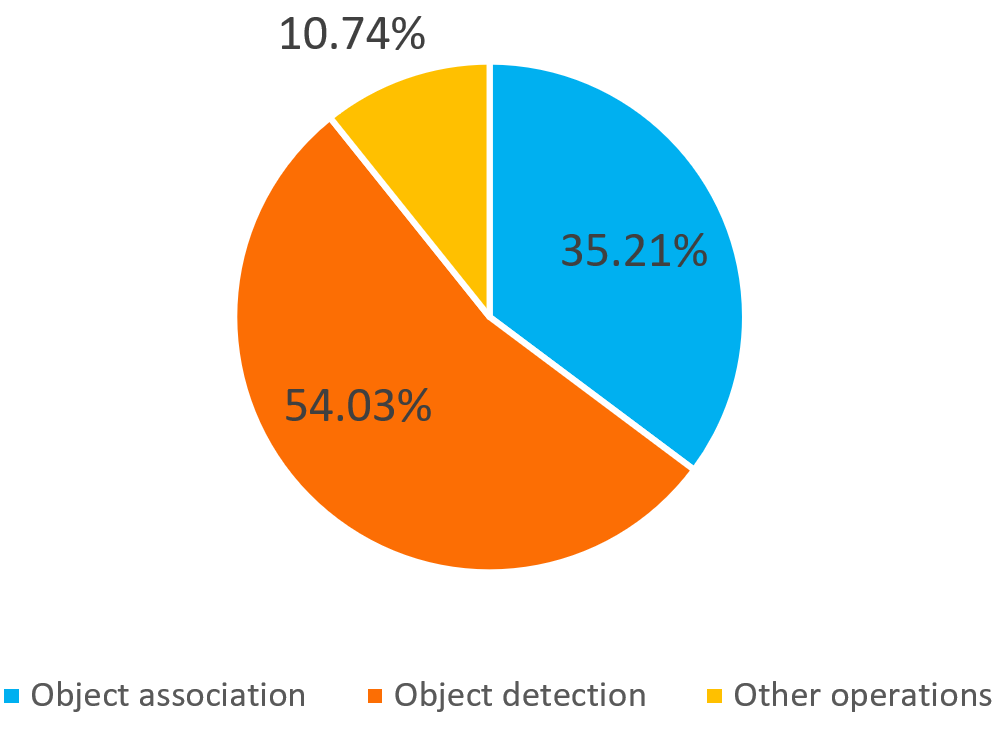}
\caption{A breakdown of GPU processing time in FairMOT on the MOT17 dataset. The frame size is fixed at $1088\times 608$ px. {Object detection} includes CPU to GPU transfer and running the DLA-34 network and generating the bounding boxes. Object association pertains to running bipartite graph matching for IOU/re-ID feature association purposes. Other operations include I/O, image resizing. }
\label{fig:timing_breakdown}
\end{figure}

\begin{figure*}[t]
\centering
\includegraphics[scale=0.23]{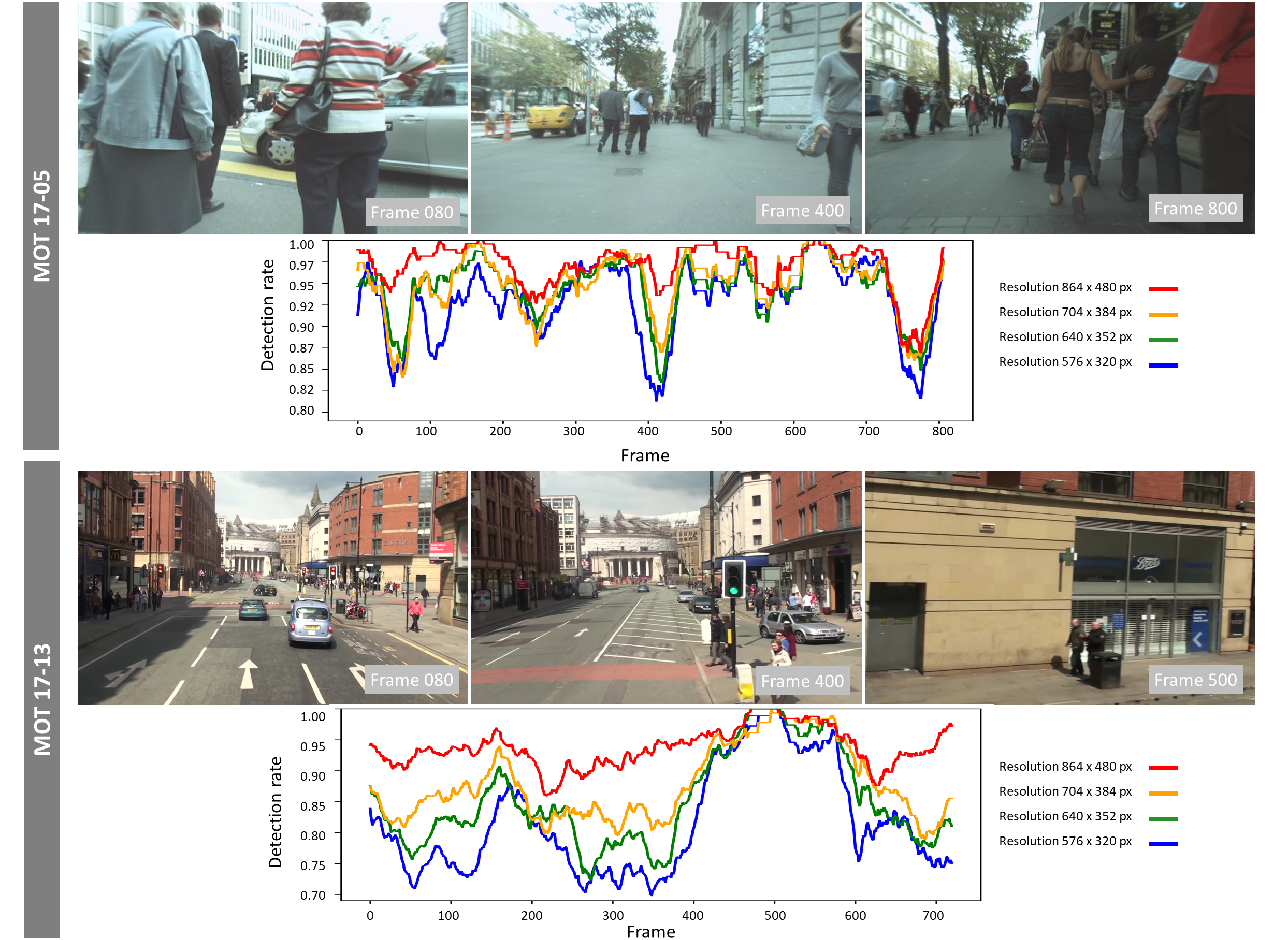}
\caption{Person detection rate using DLA-34 on MOT17 dataset. We normalized the numbers of detected objects to that in the full resolution detection.}
\label{fig:det_perf_vs_res}
\end{figure*}

To gain insights on the effect of frame sizes on tracking performance, we have conducted experiments using FairMOT~\cite{zhang2020fairmot}, a SOTA tracker on MOT datasets. The FairMOT object detector uses DLA-34~\cite{dla} as the backbone architecture and has three heads to predict a heatmap, re-ID features, and bounding box sizes for each frame. Object association is done by initializing tracklets based on the estimated boxes in the first frame and linking the detected boxes in subsequent frames to existing tracklets based on their cosine distances computed on re-ID features and the amount of bounding box overlaps. A Kalman filter is also used to predict the locations of the tracklets in the current frame.

Figure~\ref{fig:timing_breakdown} shows a breakdown of the total tracking time among object detection, object association, and other operations such as I/O, image resizing, etc. The input frame size is $1088\times 608$ pixels. As evident from the figure,  object detection is the most time-consuming step, taking more than half of the total time. The time spent on object association is also significant. Table~\ref{table:first} compares the tracking performance in MOTA and identification F1 score (IDF1), and inference time. MOTA is
defined as  
\begin{equation}
    MOTA = 1 - \frac{\sum_{t} FP_t + FN_t + IDSW_t}{\sum_{t} GT_t},
\end{equation}
where $t$ represents the frame index, $FP_t$, $FN_t$ and $IDSW_t$ are false positive, false negative, and ID switched objects and $GT$ stands for the number of ground truth bounding boxes. 
As expected, larger frame sizes generally result in higher tracking accuracy at the cost of longer inference time.
\begin{table}[t]
\centering
 \caption{The tracking time and accuracy of FairMOT on the MOT17 dataset at different input resolutions performed on one Tesla P100 GPU}
 \begin{tabular}{c c c c} 
 \hline
 Resolution (px) & MOTA(\%) & IDF1(\%) & Time \\ [0.5ex] 
 \hline
 $ 576 \times 320 $ & 58.7 & 66.8 & 1X  \\ 
 \hline
 $ 864 \times 480 $ & 68.6 & 72.8 & 1.25X  \\
 \hline
 $ 1088\times 608 $ & 70.4 & 74.1 & 1.62X  \\
 \hline
\end{tabular}

\label{table:first}
\end{table}

However, frame resolutions do not affect tracking performance uniformly across all scenes. In Figure \ref{fig:det_perf_vs_res}, we provide two sample trials of FairMOT, one from MOT17-05 and one from MOT 17-13. To quantify detection performance, we define {\it detection rate} as the number of objects detected on frames of a fixed size normalized by the number of detected objects at the full resolution. From Figure \ref{fig:det_perf_vs_res}, we observe that in both trials, a lower resolution negatively impacts detection rates. However, the gap between the detection rates of resolution $864\times 480$ px and resolution $576\times 320$ px is less pronounced in MOT 17-05 than in MOT 17-13. This can be explained by the dominance of close and bigger objects in MOT 17-05. Furthermore, in both trials, there are scenes where tracking at the lowest resolution gives good detection rates. Examples are frame 80 in MOT 17-05 and frame 500 in MOT 17-13, which contain mostly close objects. In contrast, for frame 400 in MOT 17-05, frames 80 and 400 in MOT 17-13, only resolution $864\times 480$ px gives good detection rates. Furthermore, due to the mixture of near and far objects and poor light conditions, all resolutions fare poorly on frame 800 in MOT 17-05. 

Therefore, we conclude that there is  {\it no one-size-fit-all frame size} for real-time MOT tasks. Even for a video stream from the same camera, changes occur over time in the density and sizes of objects. Therefore, an adaptive scheme based on the analysis of the visual content in scenes is needed in order to find a good trade-off between tracking accuracy and speed.

\section{Tracking with Adaptive Resolutions}
\label{multiresolutiontracking}
\subsection{Algorithmic overview}

To take advantage of the shorter inference time of smaller frame sizes, DeepScale aims to select a suitable resolution on the fly without compromising tracking performance. We assume the underlying object detection network can inherently handle inputs of different sizes. This is true for many modern detectors based on FCNs. For example, CenterNet~\cite{duan2019centernet} outputs detection maps as grid cells. Each cell is associated with the likelihood that an object falls into the cell. In absence of any fully connected layer, FCNs can produce detection maps of different sizes for different input resolutions. 
 
\begin{center}
 \begin{figure*}[]
\includegraphics[scale=0.27]{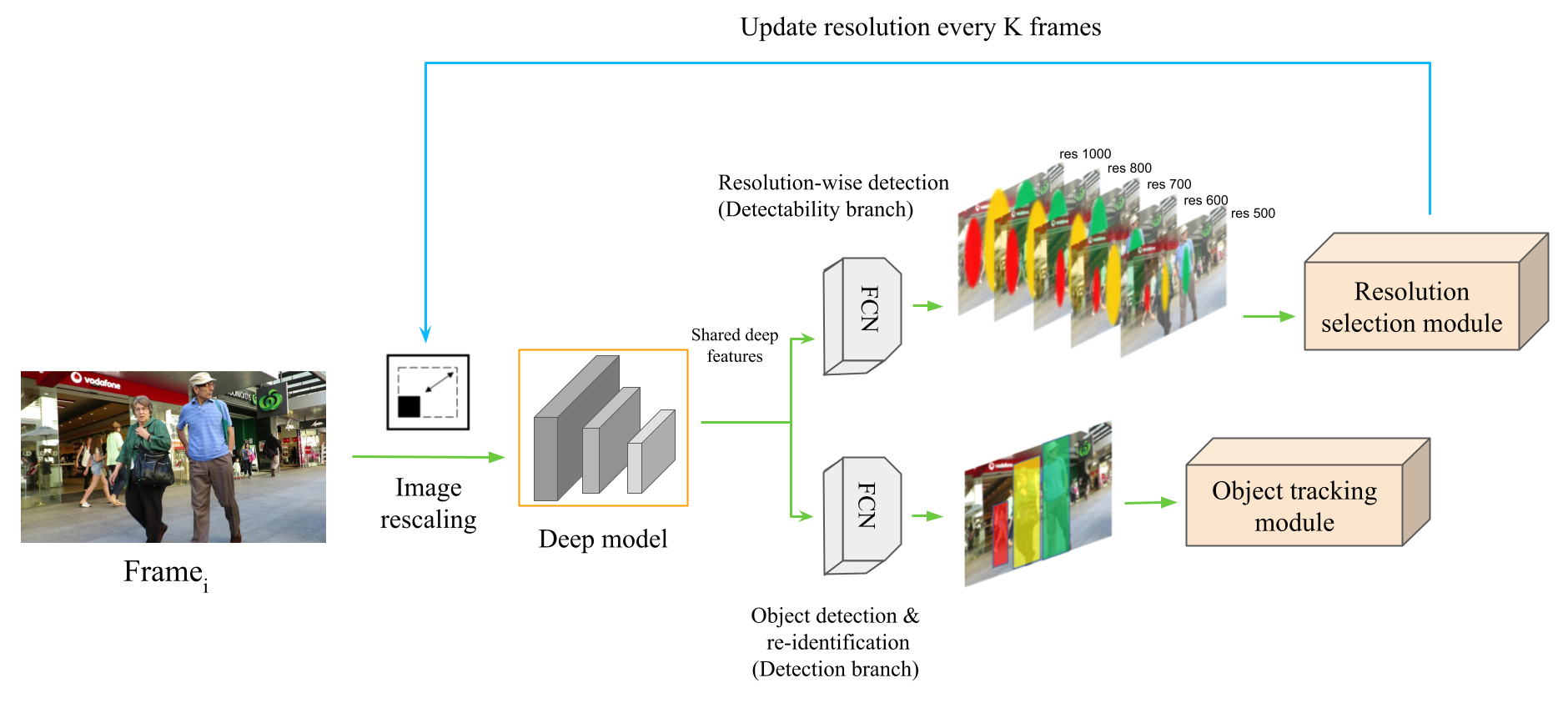}
\caption{The DeepScale pipeline - DeepScale performs object tracking and resolution estimation simultaneously via two FCN~\cite{long2015fully} heads: detection and resolution estimator. First, the received input is resized to the highest image size. Then, for every $K$ frame, DeepScale generates two outputs: detected bounding boxes and resolution-wise feature maps. Finally, resolutions for the next $K-1$ frames are updated based on the result generated from \textit{Resolution selection module},  Eq.\eqref{eq:5}.}
\label{fig:pipeline}
\end{figure*}
\end{center}

 Figure \ref{fig:pipeline} illustrates the DeepScale pipeline. DeepScale runs on every $K$th frame. Given a set of $R$ candidate resolutions $\{\beta_1, \beta_2, \ldots, \beta_R\}$, it takes a full-resolution frame of size $H\times W$ (e.g., $1088\times 608$ px) and outputs one heat map of size $H'\times W'$ for each resolution in parallel. In other words, the output is a tensor of dimension $R\times H'\times W'$. Then, these heat maps are used to estimate how well frames of the corresponding resolutions {\it preserve} the detectability of the full-resolution frame. The smallest resolution that meets the detectability requirement will be selected. The subsequent $K-1$ frames are then down-sampled to the selected resolution for tracking. 

\subsection{Self-supervised learning}
In DeepScale, we choose not to predict the ``optimal" resolution directly for two reasons. First, a single-valued resolution does not provide sufficient supervision signals to train the model. Second, since one can not simultaneously optimize tracking accuracy and inference time by scaling input frame sizes, users should be given the choice to determine the best trade-off for their applications. The DeepScale network learns the heatmaps for input frames of different resolutions. Doing so also has the added benefit of eliminating the need for labeled training data for resolution selection. 

Formally, DeepScale learns a mapping $T:  \mathbb{R}^{ 3 \times H \times W} \mapsto  [0,1]^{R\times H' \times W'}$. 
To prepare the training data for DeepScale, for each full-resolution RGB frame of size $H\times W$, ground truth heatmaps for $R$ resolutions are generated as follows. First, for each resolution $\beta_r$, we resize the training set to frames of size ${H}_{r}\times {W}_{r}$. Second, we run the base detector (e.g., FairMOT) to output the centers and corners of bounding boxes of detected objects in each frame {and scale them (up) to values in a full frame}. Third, similar to ~\cite{law2018cornernet}, {we apply a 2D Gaussian filter over each object center by penalizing the pixels outside the detected bounding boxes based on their distances to the center of the bounding box.} 
Specifically, let $(x_{c,r},y_{c,r)}$ be the centers  of a detected object in a frame of resolution $\beta_r$. Then,
$Y_{r,x,y}= exp(-\frac{{(x_r-x_{c,r})}^2+{(y_r-y_{c,r})}^2}{2\sigma^2_{c,r}})$ is the pixel value at location $(x_r, y_r)$ in the heat map for resolution $\beta_r$ augmented with a Gaussian kernel, and $\sigma_{c,r}$ is determined by the object size and the relative resolution. $Y_{r,x,y}$ acts as ground truth labels to train the detectability branch of DeepScale. Therefore, our approach is self-supervised. 

Let $\hat{Y}_{r,x,y}$ be the predicted heatmap for resolution $\beta_r$. We define a detectability loss using pixel-wise logistic regression with a focal loss as~\cite{lin2017focal}:
\begin{equation}
    \label{eq:4}
    Loss_{detectability} = -\frac{1}{N}\sum_{r,x,y} 
    \begin{cases}
     f(\hat{Y}_{r,x,y}) \quad \quad \quad \text{$Y_{r,x,y} = 1$};\\
     g(Y_{r,x,y},\hat{Y}_{r,x,y})\quad \text{otherwise};
\end{cases}
\end{equation}
where
\begin{align*}
  f(\hat{Y}_{r,x,y})&=(1-\hat{Y}_{r,x,y})^\alpha log(\hat{Y}_{r,x,y})\\
  g(Y_{r,x,y},\hat{Y}_{r,x,y})&=(1-Y_{r,x,y})^\beta (\hat{Y}_{r,x,y})^\alpha log(1-\hat{Y}_{r,x,y})
\end{align*}
and $\alpha=2$ , $\beta=4$ are focal loss hyper-parameters, and $N$ is the number of objects  in the frame.

Finally, we jointly train the detection and detectability branches using the following loss function:
\begin{equation}
    Loss = Loss_{detection} + \lambda Loss_{detectability},
\end{equation}
{where $Loss_{detection}$ pertains to the detection sub-network of the base model { as follows: } 
\begin{equation}
\begin{split}
    &Loss_{detection}  = \\ &\frac{1}{2}(\frac{1}{e^{\omega_1}}(L_{heatmap}+L_{box})+\frac{1}{e^{\omega_2}}L_{identity}+\omega_1+\omega_2),\\
    \end{split}
    \end{equation}
{which consists of learnable task based parameters $\omega_1$ and $\omega_2$, heatmap loss, box-size loss, and re-identification loss defined in  ~\cite{zhang2020fairmot}}\footnote{Heatmap loss in $Loss_{detection}$ is only for frames of the full resolution. In contrast, $Loss_{detectability}$ sums over Gaussian filtered heatmap losses over all resolutions.}. In the experiments, we set $\lambda=1$. }
\subsection{From heatmaps to tracking with adaptive resolutions} \label{pipepline_algorithm}
At the inference stage, for every $K$th frame, DeepScale predicts heatmaps $\hat{Y}_{r,x,y}, r = 1, 2, \ldots, R$, $x = 1, 2, \ldots H'$, $y = 1, 2, \ldots W'$. This part is presented as \textit{Resolution selection module} in Figure \ref{fig:pipeline}. 

The task of finding the best resolution can be  formulated as:
\begin{equation}
\label{eq:5}
\begin{aligned}
\min \quad & r\\
\textrm{s.t.} \quad & \frac{\sum^{H',W'}_{x,y}{\hat{Y}_{r,x,y}\circ \hat{Y}_{max,x,y}}}{\sum^{H',W'}_{x,y}{\hat{Y}_{max,x,y}}} \ge \gamma_r, \\
\end{aligned}
\end{equation}
{where $\hat{Y}_{max,x,y}, x = 1, 2, \ldots H'$, $y = 1, 2, \ldots W'$ is the predicted heatmap for the highest resolution,`$\circ$' is the Hadamard product operator, and  $\gamma_r\in[0,1]$ is a user-specified threshold for resolution $\beta_r$ and $\gamma_{r_i} \ge \gamma_{r_j}$ if $r_i > r_j$. $\gamma_r$ reflects users' tolerance of degradation in detectability. Clearly, smaller $\gamma_r$s favor lower resolution images. 
 In Section \ref{multi_resolution_training} we will discuss how to choose appropriate threshold values in the face of different latency-accuracy requirements.} The solution to Eq.\eqref{eq:5} gives the resolution to be used for the next $K-1$ frames. {Once the suitable resolution is found, DeepScale proceeds to construct trajectories of detections.} 

For object association, trajectories are initially created from detected objects in the first frames and extended by linking newly detected ones in subsequent frames. To link objects, we leverage appearance-based re-ID features {(of 128 dimensions in our implementation) {extracted from the re-ID branch. By sharing the same feature extraction network between the detection and detectability branches, less overhead is incurred.} We compute an appearance affinity matrix based on the cosine similarity and assign each new detection to the existing trajectories using bipartite marching.} 
If a new object fails to match an existing trajectory according to appearance, we further verify if its bounding box has significant overlap with existing ones, called {\it Intersection over Union (IoU) criterion}. Similar to ~\cite{jde} we use a Kalman filter ~\cite{welch1995introKalman} to predict the coordinates of previously detected objects in the current frame.
Note that changing input resolutions between consecutive frames may lead to different re-ID features for the same object and thus a mismatch. However, the use of the IoU criterion can mitigate such a negative effect since the movements of objects between neighboring frames tend to be small. 

{During training using Eq.\eqref{eq:4}, DeepScale learns the quality of detection for each resolution offline. Given a user's accuracy-latency requirement at run time, it selects a set of suitable thresholds and adapts the input size by solving Eq.\eqref{eq:5} online. No additional training is needed for different threshold values. To ease user selection, we also provide three set configurations (in Section~\ref{adaptive_vs_fixed}, Table \ref{table:thresh_choose_3}) that correspond to high-accuracy \& low-speed, medium accuracy \& medium speed, and low-accuracy \& high-speed, respectively.}
%
\subsection{Multi-resolution training}
\label{multi_resolution_training}
Due to the use of FCNs, DeepScale can handle different frame sizes during inference. We can take advantage of such property during training as well by augmenting the training data with five input frame sizes via resizing. Furthermore, the network meets all five image resolutions during each training epoch. Since the network structure and parameters remain the same, switching from one resolution to another does not impose any extra overhead. We call this extension {\it DeepScale++}. DeepScale++ differs from DeepScale only in the training phase. During inference, the same process is followed by both models.
\section{Computation Partition for MOT on Smart Camera-Edge with DeepScale} 
\label{sect:partition}
\begin{center}
 \begin{figure*}[]
\includegraphics[scale=0.14]{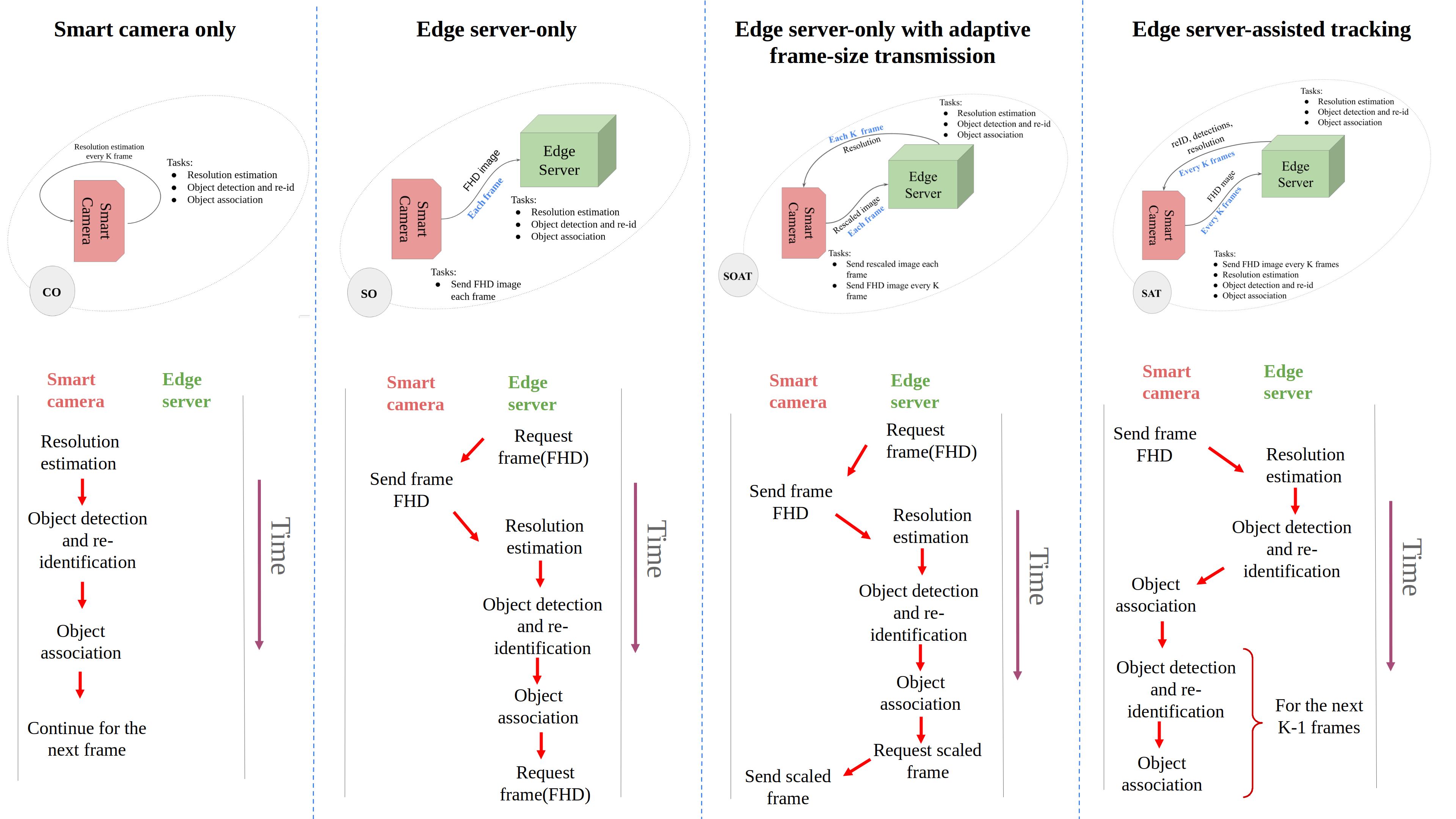}
\caption{Four representative architectures for computation partition between an edge server and a smart camera. From left to right: Smart camera-only: the end device runs DeepScale (adopting a lightweight object detection model e.g. YOLO) locally; Edge server-only: the edge server receives FHD frames and performs tracking using the DLA-34 model; Edge server-only with adaptive frame-size transmission: same as SO but the edge server adjusts the image resolution sent by the smart camera; Edge server-assisted tracking: the smart camera runs DeepScale and each $K$ frame receives the optimum resolution and detection results from the edge server.}
\label{fig:client_server_rel}
\end{figure*}
\end{center}

During the interference phase, DeepScale can adaptively select frame resolutions based on the user-specified quality of service requirements. When deployed on smart cameras, depending on their computation capability and access bandwidth, the DeepScale pipeline as shown in Figure~\ref{fig:client_server_rel} can be flexibly partitioned between smart cameras and edge servers (shortened as {\it server} in the following discussion). Specifically, we consider four representative architectures:
\begin{itemize}
    \item{\bf Smart camera-only (CO):} All computation is done on an edge device. DeepScale is employed to accelerate the processing time on the edge.
    \item{\bf Edge server-only (SO):} Full-resolution frames are sent to a server, where all computation is performed. DeepScale is employed to accelerate the processing time on the server. 
    \item{\bf Edge server-only with adaptive frame-size transmission (SOAT):} For every $K$th frame, a full-resolution (FHD) frame is sent to a server, which in turn informs the edge device of the suitable frame solution for the subsequent $K-1$ frames to send. All computation is done on the server. DeepScale is employed to accelerate the processing time on the server. 
    \item{\bf Edge server-assisted tracking (SAT):} For every $K$th frame, a full-resolution frame is sent to a server, which in turn computes the bounding boxes and reID features, and determines the suitable resolution for subsequent frames. The results are sent to the edge device.  Upon reception of the information, the edge device determines object association. Additionally, the edge device performs object detection and association for the remaining $K-1$ scaled frames. In this setup, DeepScale partly runs on the server and partly on the edge device. 
\end{itemize}

Clearly, there exist trade-offs among computation on the smart camera and the edge server and the amount of network data transfer. A qualitative comparison is given in Table~\ref{table:q_comp_edge_server_arch }. Quantitative experimental results from a real-world testbed can be found in Section~\ref{testbed_exps}. 

In addition to the aforementioned architectures and control parameters (e.g., $K$), further trade-offs can be made by running different backbone models thanks to the model agnostic nature of DeepScale. This is particularly relevant in the SO and SOAT, where all or most of the computation is done on an edge server. In such situations, a lower-capacity backbone model can be run on the edge device for feature extraction likely at the cost of degraded tracking accuracy.

\begin{table}[t]
    \centering
      \caption{Qualitative Comparison Among Different Edge-Server Architectures}
    \begin{tabular}{c|c|c|c|c}
    \hline
         & \multicolumn{2}{c|}{Computation} & \multicolumn{2}{c}{Network Traffic} \\ \hline
       Approaches & Server & Edge & S $\rightarrow$ E & E $\rightarrow$ S \\ \hline \hline
       CO  & No & High & No & No\\ \hline
       SO & High & No & No & High\\ \hline
       SOAT & High & No & Low & Low\\ \hline
       SAT & Low & High & Low & Low\\ \hline
    \end{tabular}
    \label{table:q_comp_edge_server_arch }
\end{table}

\section{Experiments}\label{experiments}
In this section, we evaluate the performance of  DeepScale on MOT datasets and a small-scale testbed focusing on pedestrian tracking tasks.  

\subsection{Implementation}
DeepScale can work with {\it any FCN-based object detection models}. In the implementation, we use FairMOT~\cite{zhang2020fairmot} and append a detectability branch consisting of two fully convolutional layers. {Among backbone networks such as ResNet-34~\cite{he2016resnet}, ResNet50~\cite{he2016resnet}, High-resolution Network(HRNet)~\cite{wang2020hrnet}  and DLA-34~\cite{dla}, from the reported tracking performance and tracking latency in ~\cite{zhang2020fairmot}, we find that DLA-34 offers the best trade-off.} However, it should be noted that DeepScale can work with the other architectures as well. {In the case that a non-FCN object detector is used, DeepScale can still be applied by storing one object detector network for each input size and performing switches based on the resolution selected during inference.} The candidate resolutions are $\{576\times 320, 640 \times 352, 704 \times 384, 864 \times 480, 1088 \times 608\}$ px.
We pre-train DeepScale on the CrowdHuman ~\cite{shao2018crowdhuman} dataset. The Adam optimizer, learning rate $e^{-4}$ and a batch size of 12 are used for 30 epochs on the training set. We augment the training datasets using {random affine transforms with parameters $scale=[0.50, 1.20]$, $rotation=[-5, 5]$, and $translation=[0.10, 0.10]$}. Testing and training are done on Tesla-P100 GPUs.

\paragraph*{Dataset}{ We use the MOT17~\cite{milan2016mot16} dataset for training, validation, and testing. The MOT17 training set includes 7 different pedestrian tracking sequences with frame resolution $[1920\times1080]$ and $[640\times480]$ pixels. The MOT17 test set consists of the same number of sequences containing different crowd scenes.}
\paragraph*{Metrics} To evaluate the tracking performance, the MOTA score is utilized.  In benchmark experiments, we also evaluate Identity Switches (IDSw), Frames per Second (FPS), Most Tracked (MT) ratio for $> 80$\% cases, and Most Lost (ML) ratio for $<20$\% cases and report IDF1 in overall tracking accuracy.

\subsection{Evaluation using MOT datasets}
\label{multi_resolution_training}

\begin{figure}[t]
\centering

\includegraphics[width=0.45\textwidth]{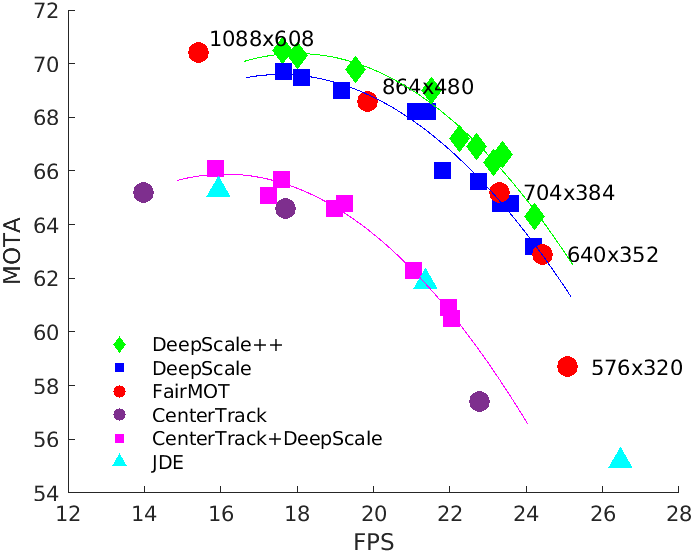}
\caption{Time efficiency of DeepScale and DeepScale++ in comparison to the fixed resolutions on the validation set of MOT17 dataset.}
\label{fig:multi_res_training_figure}
\end{figure}

In Figure \ref{fig:multi_res_training_figure}, we compare the tracking performance and speed of DeepScale, DeepScale++ and several SOTA methods. We use a pre-trained model for CenterTrack and train both FairMOT and JDE on half of the  MOT17 training set. FairMOT and CenterTrack use {DLA-34 as their detector's backbone and DarkNet-53~\cite{yolov3} is the network architecture for the JDE tracker.} All of these three trackers can only handle frames of a fixed set of resolutions. In contrast, in DeepScale and DeepScale++, by changing configuration parameters (e.g., thresholds), different trade-offs between MOTA and FPS can be achieved. The threshold settings for DeepScale++ in the experiments are given in Table~\ref{table:thresh_choose_2}, where the minus sign means the corresponding resolution cannot be chosen. 
In the figure, {2nd order} polynomial fitting functions for the results of DeepScale and DeepScale++ are also plotted. 

It is worth noting that the $\langle$MOTA, FPS$\rangle$ tuples achieved by FairTracker under different input frame sizes generally fall on the curve associated with DeepScale. This indicates that DeepScale does not compromise tracking performance while providing users the flexibility in choosing proper trade-offs between tracking accuracy and speed. From Figure \ref{fig:multi_res_training_figure}, we also observe that the DeepScale++  curve lies on the top and to the right of that of DeepScale. This implies that for the same tracking speed, DeepScale++ can achieve higher MOTA. Conversely, for the same MOTA, DeepScale++ takes less time than DeepScale. Compared to FairMOT on full-resolution frames, the MOTA score of one configuration of DeepScale++ is higher and its frame rate is 14\% faster. 

{We also study the application of DeepScale on CenterTrack. During inference, CenterTrack dynamically switches the input frame sizes according to the selected resolutions from the DeepScale pipeline. Compared to CenterTrack with fixed resolution, the "CenterTrack+DeepScale" curve indicates faster performance and more accurate tracking results.}
\begin{table}[t]
\centering
\caption{The tracking accuracy and latency of DeepScale++ under different thresholds, $K = 40$}
\resizebox{\columnwidth}{!}{%
\renewcommand{\arraystretch}{1.00} 
 \begin{tabular}{ c|c|c|c|c|c }
 \hline
 \multicolumn{4}{c|}{Threshold}&\multicolumn{1}{c|}{\multirow{2}{*}{\makecell{MOTA}}} & \multicolumn{1}{c}{\multirow{2}{*}{\makecell{FPS}}}\\ 
 \cline{1-4}
  $864\times480$ & $704 \times 384$ &$640\times 352$& $576\times 320$&&\\ \hline
  0.00&0.00&0.00&0.80&62.2&24.53\\\hline
  0.00&0.00&0.00&0.90&64.2&24.38\\\hline
  0.00&0.00&0.00&1.00&64.3&24.33\\\hline
  0.00&0.00&0.80&0.90&64.6&24.14\\\hline
  0.00&0.00&1.00&1.00&66.3&23.69\\\hline
  0.80&1.00&1.00&1.00&69.0&21.62\\\hline
  0.90&1.00&1.00&1.00&69.8&21.31\\\hline
  1.00&1.00&1.00&1.00&69.8&19.63\\\hline
  0.90&\textendash&\textendash&\textendash&70.4&19.32\\\hline
  1.00&\textendash&\textendash&\textendash&70.8&17.63\\\hline
\end{tabular}}
\label{table:thresh_choose_2}
\end{table}

\paragraph*{Adaptive vs fixed resolution tracking} \label{adaptive_vs_fixed}

We demonstrate the ability of DeepScale to tune trade-offs between tracking accuracy and speed for different parameter settings and compare the results against those from fixed resolutions. In the experiments, for simplicity, only three threshold settings are considered as listed in Table \ref{table:thresh_choose_3}. These configurations are chosen to represent, respectively, low latency-low accuracy (C1), medium latency-medium accuracy (C2), high latency-high accuracy (C3). 
\begin{table}[h]
\centering
\caption{Three set configurations for DeepScale and DeepScale++}
\resizebox{\columnwidth}{!}{%
\begin{tabular}{l|c|c|c|c}
    \hline 
     Size & $864\times 480$ & $704 \times 384$ & $640\times 352$ & $576\times 320$ \\
     \hline
     Threshold & $\gamma_1$ & $\gamma_2$ & $\gamma_3$ & $\gamma_4$ \\
    \hline
    C1 & 0.00 & 0.00 & 1.00 & 1.00 \\
    C2 & 0.80 & 1.00 & 1.00 & 1.00 \\
    C3 &  1.00 & \textendash & \textendash  &\textendash\\
    \hline
\end{tabular}}
\label{table:thresh_choose_3}
\end{table}

Table \ref{table:low_resolution_afford} summarizes the tracking performance and speed for DeepScale and from fixed resolution inputs. {From the table, we observe that  DeepScale++C3 outperforms tracking with full-resolution frames in MOTA and is 14\% faster. Similar improvements are also achieved by DeepScale++C2 and DeepScale++C1 relative to the respective fixed-size inputs.}

\begin{figure}[t]
\centering
\includegraphics[scale=0.30]{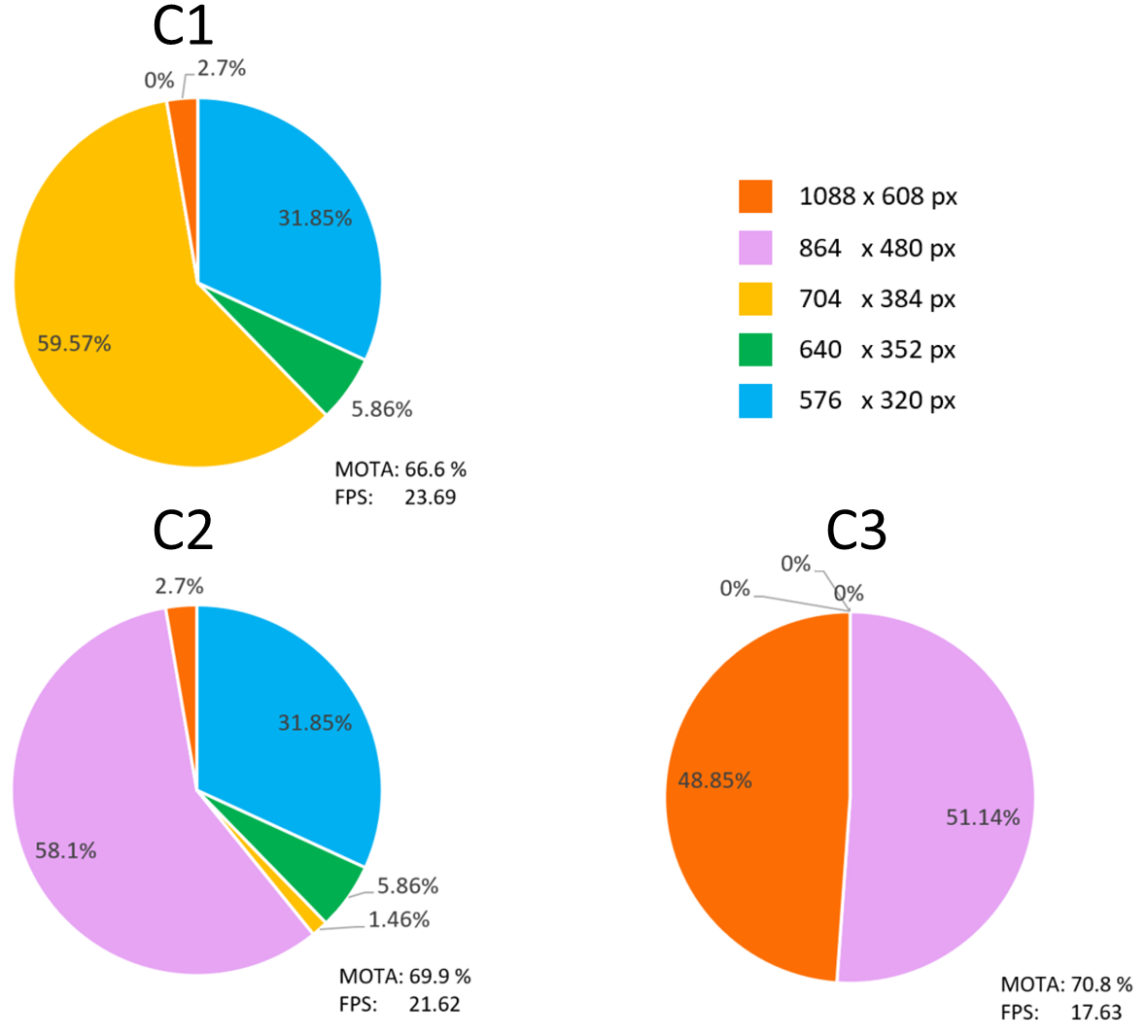}
\caption{Percentages of frames of different resolutions selected by DeepScale++ under different configurations on the validation set of MOT17, $K = 40$.}
\label{fig:resolution_breakdown}
\end{figure}

We further show in Figure~\ref{fig:resolution_breakdown} the percentage of frames of different resolutions selected by DeepScale++ under different configurations. With DeepScale++C3, 48.85\% of frames are processed at high resolution, which explains its high MOTA score. In contrast, DeepScale++C2 processes very few (2.7\%) high resolution frames and significantly more second highest ($704\times 384$ px) and a significant percentage (31.85\%) of lowest-resolution frames ($576\times 320$ px). The trend continues with DeepScale++C1, which selects the 3rd highest resolution for 59.57\% of input frames and the lowest resolution for 31.85\% of frames. 
\begin{table}[t]
\centering
\renewcommand{\arraystretch}{1.00} 
\caption{Adaptive Frame Size vs Fixed Size Tracking on the validation set of MOT17 dataset, $K = 40$}
\begin{tabular}{lll}
\hline
 Fixed size or DeepScale++ & MOTA\% & FPS\\ \hline
 $1088\times 608$ px & 70.4 & 15.43 \\ 
 DeepScale++C3 & 70.8 & 17.63 \\ 
 $864\times 480$ px & 68.6 & 19.85 \\ 
 DeepScale++C2 & 69.0 & 21.62 \\ 
 $704\times 384$ px & 65.2 & 23.30 \\ 
 DeepScale++C1 & 66.3 & 23.69 \\ 
 $640\times 352$ px & 62.9 & 24.43 \\ \hline
\end{tabular}

\label{table:low_resolution_afford}
\end{table}

\paragraph*{Impact of adaptation interval $K$}
Next, we investigate the impact of $K$ on tracking performance and speed. Recall that DeepScale is applied every $K$ frames to determine the suitable resolutions for the next $K-1$ frames. When $K$ is smaller, resolution selection is done more frequently and thus leads to better tracking performance at the cost of more computation overhead. Table~\ref{table:table_samplingRate} summarizes the performance of DeepScale under three configurations over 4 different adaptation intervals. {In general, as $K$ decreases, tracking speed reduces but tracking accuracy improves. Among the three configurations, the value of $K$ has the most impact on DeepScale++C1 with MOTA scores increasing from 66.3\% to 69.3\% when $K$ reduces from 40 to 2. Somewhat surprisingly, with DeepScale++C3, very few changes are observed in MOTA scores for different Ks. This may be explained by multi-resolution training in DeepScale++, which makes object detection more robust to smaller object sizes in lower-resolution frames}. Indeed, we observe from experiments with DeepScale (omitted due to space limits)  consistent improvements in MOTA scores for smaller $K$ among all configurations.  
\begin{table}[t]
\centering
\caption{Impact of Adaption Interval $K$ on the Validation Set of MOT17 Dataset}
\resizebox{\columnwidth}{!}{%
 \begin{tabular}{c|c|c|c|c|c|c|c|c}
 \hline
 \multicolumn{1}{c|}{\multirow{2}{*}{\makecell{DeepScale++ \\configuration}}}&\multicolumn{2}{c|}{K=40}&\multicolumn{2}{c|}{K=20}&\multicolumn{2}{c|}{K=10}&\multicolumn{2}{c}{K=2}\\
 \cline{2-9}
  &FPS&MOTA&FPS&MOTA&FPS&MOTA&FPS&MOTA\\
 \hline
 C3&17.63&\textbf{70.8}&17.61&70.6&17.44&70.7&16.52&\textbf{70.7}\\
 C2&21.62&69.0&21.35&69.5&20.87&69.6&18.37&70.4\\
 C1&\textbf{23.69}&66.3&23.44&66.6&22.94&66.7&19.58&69.3\\
 \hline
\end{tabular}}

\label{table:table_samplingRate}
\end{table}
\begin{table*}[t]
\centering
\caption{Results of four representative architectures for computation partition between edge server and smart cameras. K=5}
\begin{adjustbox}{width=1\textwidth}
\begin{tabular}{c|c|c|c|c|c|c|c}
    \hline 
     Architecture & \makecell{DeepScale++\\configuration}& FPS$\uparrow$ & MOTA(\%)$\uparrow$ & \makecell{Server time$\downarrow$\\(ms) per frame} & \makecell{Camera time$\downarrow$\\(ms) per frame} & \makecell{Transmission time$\downarrow$\\(ms) per frame}&\makecell{Network traffic load$\downarrow$\\KB per frame} \\
     \hline
     SO&C1&5.7&66.3&43.6&55.6&87.4&46.5\\
     SO&C2&5.6&\textbf{68.7}&46.8&53.2&75.7&46.5\\
     SO&C3&5.2&\textbf{69.5}&57.1&52.8&80.4&46.5\\
     SOAT&C1&\textbf{7.0}&66.2&41.9&34.8&67.7&26.9\\
     SOAT&C2&\textbf{6.4}&\textbf{68.5}&46.2&38.6&69.2&31.8\\
     SOAT&C3&5.5&\textbf{69.7}&56.8&46.3&76.9&40.8\\
     SAT&C1&5.0&56.8&17.9&158.0&22.1&21.6\\
     SAT&C2&4.5&60.6&19.5&179.3&22.0&21.6\\
     SAT&C3&3.8&61.9&17.8&223.8&22.2&21.6\\
     \hline
     CO&C1&5.3&55.9&0.0&186.0&0.0&0.0\\
     CO&C2&4.7&59.9&0.0&121.6&0.0&0.0\\
     CO&C3&3.7&61.6&0.0&267.5&0.0&0.0\\
    \hline
\end{tabular}
\end{adjustbox}
\label{table:camera_server_exps}
\end{table*}

More experiments of DeepScale++ and baseline models on MOT benchmarks can be found in Section \ref{sect:app}.

\subsection{Testbed Experiments}
\label{testbed_exps}
\paragraph*{Testbed} {Our testbed consists of an embedded device mimicking a smart camera and a server. Specifically, the end device is NVIDIA Jetson TX2 which adopts an integrated GPU. The edge server has a single Tesla-P100 GPU, Intel(R) Xeon(R) CPU, and 64GB RAM. The end device connects to the server through a Wi-Fi router. The network upload and download bandwidths are 21.1 Mbps and 78.0 Mbps respectively, and the average round trip latency is 1.40 ms.}

\paragraph*{Implementation} {All models are implemented in PyTorch ~\cite{NEURIPS2019_9015} a popular deep learning framework in Python. Due to the resource constraints on the end device, we train a lightweight DeepScale++ model based on the YOLO\cite{yolov3} architecture. Therefore, during inference, different DeepScale models run on the end device and the edge server in SOAT. All models are trained on half of the MOT17-training dataset and validated using another remaining half. To reduce the network load, we compress the frames using image encoding implementations from OpenCV~\cite{opencv_library} library before sending them through the network.}

\paragraph*{Computation partition strategies}{ Table \ref{table:camera_server_exps} compares the accuracy and time of running DeepScale++ under different computation partition strategies in the testbed. The time spent on the end device and service include computation time while the transmission time includes the amount of time to transfer uplink (camera-server) and download (server-camera) traffics if applicable in each frame. In the experiments, $K$ is set to 5 frames. As expected, the server time in SO and SOAT are comparable and higher than that in SAT and CO.  The time spent on the end device is the opposite. When C1 (low latency-low accuracy) and C2 (medium latency-medium accuracy) are chosen, the transmission time is significantly reduced when comparing SOAT to SO. SAT has significantly lower transmission time than SO and SOAT since in SAT, FHD frames are sent only every $K$ frame to the server, which is consistent with the amount of network traffic in Table~\ref{table:camera_server_exps}. Note that CO achieves comparable FPS as SAT. But this comes at the cost of lower MOTA scores due to the use of a lightweight object detection architecture.}

\begin{figure}[t]
\centering

\subfloat[SAT: Edge  server-assisted  tracking ]{\includegraphics[scale=0.70]{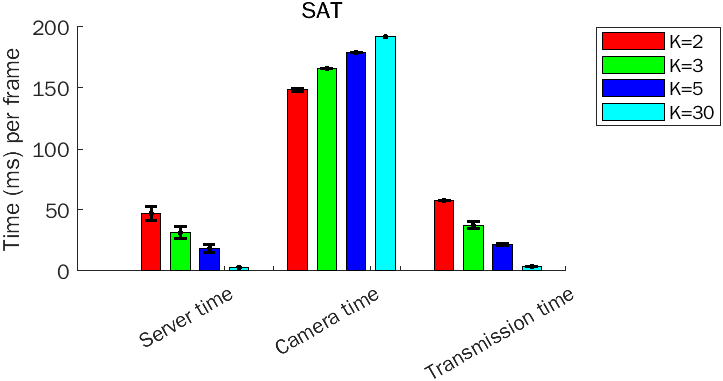}}\\
\subfloat[SOAT: Edge server-only with adaptive  frame-size transmission]{\includegraphics[scale=0.70]{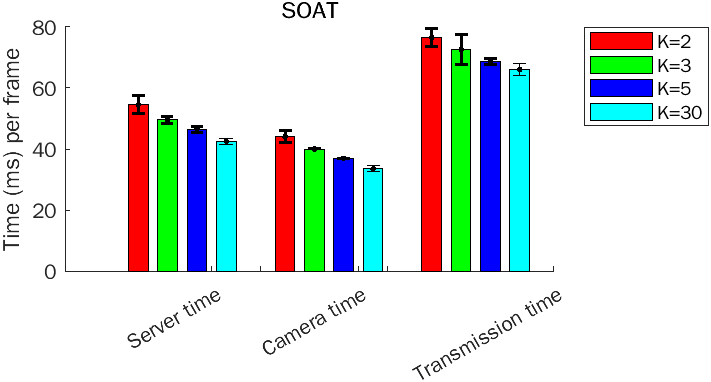}}%
\caption{Impact of interval K on workload partition}
\label{fig:various_k_in_client_server}
\end{figure}

\paragraph*{Effects of K} {Since the smart camera periodically sends full-resolution frames to the edge server in SOAT and SAT, the total amount of time spent for each frame depends on the value of $K$ in both strategies. Figure \ref{fig:various_k_in_client_server} illustrates the time spent on the edge server, the smart camera, and in data transmission between the two for SAT and SOAT. Similar to Table~\ref{table:camera_server_exps}, for different $K$s, more time is spent on the edge server and data transmission in SOAT than SAT. As $K$ increases, both server time and transmission time decrease significantly with SOAT. In comparison, the decrements in SAT are more pronounced since as $K$ increases, the smart camera processes fewer FHD frames. In summary, we observe non-trivial trade-offs among time spent on the smart camera, the server, and data transmission across different computation partition strategies and choices of configuration parameters (e.g., $K$ and C1 -- C3). There is no one-size-fits-all solution. To select the optimal strategy and parameters, one needs to profile the compute time, I/O time, and available network bandwidth in a target deployment environment.}

\section{Related Work}\label{relatedWork}
In this section, we first describe techniques for adapting frame sizes in visual processing pipelines. Next, we discuss approaches toward real-time video analytics.

\subsection{Adaptive frame size in visual processing}
The need to adapt frame sizes has been recognized in the literature on action recognition, video object detection, and classification applications to speed up the inference time. 

AdaScale~\cite{chin2019adascale} uses a scale regression model to predict a resolution that results in the minimum loss of predicted bounding boxes. The authors define a metric associated with the loss of predicted bounding boxes in an image at different scales and use it to generate the optimal scale labels by running a pre-trained object detector on the training set. A scale regressor is then trained along with a bounding box predictor in a two-headed network.  While AdaScale can be utilized in MOT, the metric proposed for determining the optimal scale fails to account for the effect of input frame sizes on re-identification and will thus lead to suboptimal performance in MOT. Moreover, AdaScale does not provide user-controlled parameters to trade-off inference performance with speed at run time. 

To meet the accuracy-latency requirements in visual object detection on embedded systems, in ApproxNet ~\cite{xu2020approxdet}, the authors introduce a data-driven modeling approach in the face of changing content and resource contention. A quadratic regression is used to pick a particular configuration at the run time among object tracking types, the number of object proposals, downsampling ratios of the tracker, and frame sizes. To select the appropriate parameters, profiling of CPU and memory usage of the configurations on a target platform is needed. As the search space grows exponentially with the number of knobs, the run time complexity increases accordingly. The use of a multi-branch detection network also increases the storage complexity of the model.

In AR-Net ~\cite{meng2020arnet}, an adaptive image-size strategy is proposed, which uses a policy network to decide what resolution to choose for action recognition at run time. The policy network contains a feature extractor and an LSTM module. To handle different input sizes, the authors propose to store different backbone networks corresponding to different resolutions. AR-NET is not directly applicable to MOT. Furthermore, it has several drawbacks: first, storing multiple networks demands extra storage, which may limit its applicability on embedded devices; second, having a separate feature extractor for the sole purpose of predicting suitable resolutions causes additional computational overhead; third, the policy network takes images of the lowest resolution as inputs. Such low-resolution images may not contain the necessary information for correct predictions (``how does one know what is not there?")

Compared to the aforementioned works, DeepScale has several advantages for MOT tasks. First, the metric to determine the optimal resolution is tailored to MOT in that it considers both target detection and re-identification across frames. Second, in DeepScale, feature extraction is shared between resolution prediction and object detection, and thus incurs small overhead. Lastly, DeepScale allows users to choose suitable accuracy-latency trade-offs at run time without having to retrain the network or store extra models/branches. 
\subsection{Real-time multi-object tracking}
Due to the high complexity of object detection and association in tracking-by-detection frameworks, existing attempts on real-time MOT fall into three categories: efficient object detection, low-latency object association, and integrated tracking models.

\paragraph*{Efficient object detection} In ~\cite{sort}, the authors employ a Faster-RCNN detection model that uses shared parameters for region proposal and object classification. In ~\cite{xu2018real}, to enable real-time MOT on unmanned aerial vehicles, the authors utilize a YOLO object detector and JPDA multiple object tracking. Although a frame rate of 60Hz can be achieved using NVIDIA Jetson TX2 on PETS datasets, more than 20\% reduction in MOTA is reported in~\cite{xu2018real}. 

\paragraph*{Low latency object association}  Methods in this category aim to accelerate tracking speed by incorporating fast and efficient data association models. The work in ~\cite{choi2015near} achieves near-real-time tracking performance through hypothesis generation and selection. In ~\cite{kim2021discriminative},  a data association module based on an LSTM network has been implemented that simultaneously considers the appearances of each track to match with detected objects in the next frame. 
\paragraph*{Integrated tracking models} Recently, several works perform object detection, compute visual features for re-identification (re-ID) and/or predict associations among objects using a single network(~\cite{zhang2020fairmot},~\cite{centerTrack},~\cite{jde}). These methods are appealing due to less inference time and higher accuracy compared to their counterparts that only consider temporal association.  In CenterTrack~\cite{centerTrack}, detected objects are associated  through time using 2D displacement prediction. In JDE \cite{jde} and FairMOT~\cite{zhang2020fairmot}, motion estimation is done using Kalman filters. In ~\cite{motsmultiobject}, the authors associate detected objects over time utilizing an extended RCNN with an association head to capture the Euclidean distance of embedding vectors.

In this work, different from existing attempts to accelerate visual processing or MOT, we adapt the sizes of input frames to achieve a good trade-off between inference time and tracking accuracy. DeepScale is model agnostic and can work with any fully convolution network (FCN)-based object detectors. It is to the best of our knowledge, the first work to do so for MOT. 

\subsection{Efficient Edge-Cloud Architecture}
Due to the limited computation capability of end devices and long network latency to transfer a large volume of data to cloud servers, efficient edge computing architectures have been investigated for visual analytics in recent years.  In ~\cite{gu2019collaborative}, Gu {\it et al.} improve the object tracking performance and energy consumption on end devices using a collaborative edge-cloud architecture where the end device offloads computation to gain more accurate object positions. In ~\cite{cao2021edge}, a difficult-case discriminator is introduced to classify images into easy and hard classes based on the extracted semantics of each image. The hard cases are uploaded to a server while easy classes are processed on the device locally. In ~\cite{ko2018edge} KO {\it et al.} propose a network partitioning solution between an edge and a host to enhance the edge platform's throughput. In the work, a DNN works as an encoding pipeline and the output of an intermediate layer is sent to the host. In ~\cite{fang2019teamnet}, Fang {\it et al.} proposed a novel distributed collaborative framework to run compute-intensive inference tasks on small specialized models executed resource-constrained devices. The key idea is to train multiple scale-down models, who together have comparable or even better inference performance than a monolithic large model. Although these approaches split computation loads between end devices and edge servers or among end devices, they are not designed for MOT and fail to account for multi-object detection quality in optimizing the end-to-end performance.

\section{Conclusion}\label{conclusion}
In this paper, we presented DeepScale, a model-agnostic method to accelerate tracking speeds for MOT tasks useful for both server and smart camera platforms, by dynamically adapting sizes of input frames at run time. It can work with any FCN-based object detection model and provide user's control knobs to determine a suitable trade-off between tracking accuracy and efficiency. An extended version called DeepScale++ that trains on multi-resolution training data was also developed. We validated both solutions on multiple MOT datasets and found that they could achieve comparable tracking accuracy as state-of-the-art methods with shorter inference time. To further improve the throughput of the DeepScale pipeline, two smart camera-edge server collaborative strategies were implemented and evaluated on a small-scale testbed. Experimental results demonstrated that the proposed computation partition approaches could improve the tracking throughput and enhance the tracking accuracy.
As future work, we are interested in exploring the use of more advanced detection backbones and accelerating object association on crowded scenes. Another venue of interest is to experiment with more advanced edge devices such as Jetson Xavier in multi-camera tracking.

{\small
\bibliographystyle{ieee_fullname}
\bibliography{egbib}
}
\appendices
\section{Benchmark evaluation}
\label{sect:app}
In this section, we evaluate the performance of DeepScale++ (C2, $K=30$) on MOT15~\cite{leal2015motchallenge}, MOT16~\cite{milan2016mot16}, MOT17~\cite{milan2016mot16} and MOT20~\cite{dendorfer2020mot20} datasets {under the private object detector category. }
For all datasets, starting from a pre-trained model on CrowdHuman~\cite{shao2018crowdhuman}, we retrain DeepScale++ on the full training data. {For comparison, we consider integrated solutions including  FairMOT, CenterTrack, and JDE trackers.} 
Note that the results are obtained by running these trackers on a single Tesla-P100 GPU.

As shown in Table~\ref{table:comp_table}, DeepScale++ ranks first in tracking speed on all datasets. For MOT15, which contains both indoor and outdoor scenes with street views, {it achieves a MOTA of 58.3 (2.3\% less than the best performing tracker FairMOT on this task) at 25.87 FPS (1.57X faster than FairMOT). Compared to JDE, DeepScale++ is $\sim $1.30X faster with 7.2\% higher in MOTA on MOT16.} 
No method manages to achieve real-time tracking ($\ge 30$ FPS) on MOT20 {due to its dense crowd scenes that take a longer time for person association.} Significant speed-up is still obtained by DeepScale++ with acceptable tracking accuracy. 

\begin{table}[h]
\centering
\caption{Comparison with SOTA methods on MOT benchmarks}
\resizebox{\columnwidth}{!}{%
\begin{tabular}{llllllll}
\hline
 Dataset & Method & MOTA$\uparrow$ & FPS$\uparrow$ & IDF1$\uparrow$ & MT$\uparrow$ & ML$\downarrow$ & ID Sw$\downarrow$ \\\hline 
\multirow{2}{*}{MOT15}  
 & FairMOT & 60.6 & 16.44 & 65.7 & 47.6\% & 11.0\% & 591\\ 
 & DeepScale++ & 58.3 & \textbf{25.87} & 62.0 & 36.3\% & 18.0\% & 572  \\ \hline
\multirow{3}{*}{MOT16}  
& JDE & 64.4 & 15.94 & 55.8 & 35.4\%& 20.0\% & 1544  \\
 & FairMOT & 74.9 & 15.43 & 72.8 & 44.7\% & 15.9\% & 1074\\ 
 & DeepScale++& 71.6 & \textbf{20.15} & 71.1 & 38.6\% & 18.7\% & 1331 \\ \hline
\multirow{3}{*}{MOT17}  
& CenterTrack & 67.8 & 13.97 & 64.7 & 34.9\% & 24.8\% & 2898\\
 & FairMOT & 73.7 & 15.43 & 72.3 & 43.2\% & 17.3\% & 3303\\ 
 & DeepScale++ & 70.2 &\textbf{20.50} & 70.3 & 39.2\% & 20.3\% & 3870 \\ \hline
\multirow{2}{*}{MOT20}
 & FairMOT & 61.8 & 12.5 & 67.3 & 68.8\% & 7.6\% & 5243\\ 
 & DeepScale++  & 59.44 &\textbf{14.39} & 66.33 & 52.6\% & 11.2\% & 4123 \\ \hline
\end{tabular}}
\label{table:comp_table}
\end{table}
\end{document}